\documentclass[conference]{IEEEtran}
\IEEEoverridecommandlockouts

\usepackage{cite}
\usepackage{amsmath,amssymb,amsfonts}
\usepackage{algorithmic}
\usepackage{graphicx}
\usepackage{textcomp}
\usepackage{xcolor}
\usepackage{url}
\usepackage[normalem]{ulem}
\usepackage{todonotes}
\usepackage{booktabs,multirow,makecell}
\def\BibTeX{{\rm B\kern-.05em{\sc i\kern-.025em b}\kern-.08em
    T\kern-.1667em\lower.7ex\hbox{E}\kern-.125emX}}

\usepackage{subcaption}

\begin{document}

\title{Symmetry-Aware Surrogate-Assisted Testing of UAVs}

\author{\IEEEauthorblockN{1\textsuperscript{st} Anas Shrinah}
\IEEEauthorblockA{\textit{Department of Software Engineering} \\
\textit{Applied Science Private University}\\
Amman, Jordan \\
a\_shrinah@asu.edu.jo}
\and
\IEEEauthorblockN{2\textsuperscript{nd} Chris Bennett}
\IEEEauthorblockA{\textit{Department of Computer Science} \\
\textit{University of Bristol}\\
Bristol, United Kingdom \\
christopher.bennett@bristol.ac.uk}
\and
\IEEEauthorblockN{3\textsuperscript{rd} Kerstin Eder}
\IEEEauthorblockA{\textit{Department of Computer Science} \\
\textit{University of Bristol}\\
Bristol, United Kingdom \\
Kerstin.Eder@bristol.ac.uk}
}

\maketitle

\begin{abstract}

We investigate a symmetry-aware surrogate-assisted testing approach for UAV autopilots. The proposed approach exploits geometric symmetries between flight missions by transforming obstacle configurations into a shared canonical coordinate frame.
This enables a neural-network surrogate trained on a single reference mission to be reused across arbitrary missions, reducing the amount of high-fidelity simulation data required for training. 
The surrogate model is trained to predict the minimum clearance distance between a UAV and surrounding obstacles and is used to guide test generation towards safety-critical scenarios. 
We evaluate this approach using the Aerialist UAV test bench and benchmark its performance against techniques from the 2024 to 2026 UAV Testing Competitions.
Results show that the approach achieves an average critical-test-case hit rate of 42\% across the evaluated missions, compared with approximately 10\% for the constrained-random baseline used to generate the training data.
This approach also achieves high input diversity and competitive benchmark performance. 
However, our findings reveal limitations in transferring surrogate predictions across missions with differing flight-path characteristics, highlighting the boundaries of geometric canonicalization.
To support reproducibility and future research, we provide the trained surrogate model and a dataset comprising more than 2,500 high-fidelity simulations.

\end{abstract}


\section{Introduction}
Unmanned Aerial Vehicles (UAVs) are increasingly employed in
applications that require autonomous navigation in complex
environments, such as indoor facilities~\cite{lam2025minav},
urban areas~\cite{lin2021urban}, and disaster
sites~\cite{oh2021disaster}.
In such environments, failures in obstacle avoidance can lead to collisions, making the safety and reliability of autonomous navigation systems a critical concern. 

To address these concerns, verification and validation methods, such as simulation-based testing, play an important role in establishing confidence in the trustworthiness of these systems. 
Simulation-based testing uses high-fidelity simulators to evaluate autonomous UAVs under a diverse range of environmental conditions while reducing the time, cost, and safety risks associated with physical flight testing.
However, the input space makes exhaustive testing impractical. Search-based testing offers a systematic approach to this challenge by guiding test generation towards scenarios that are more likely to reveal system failures~\cite{nejati2023reflections}.
This process, however, requires evaluating numerous candidate scenarios, making the computational cost of high-fidelity simulations a significant bottleneck.

Surrogate-assisted testing offers an approach to mitigating this bottleneck by using computationally inexpensive surrogate models to estimate test outcomes without executing the high-fidelity simulator~\cite{beglerovic2017testing}. 
These models learn an approximation of the relationship between test inputs and simulation outcomes using data from previously executed high-fidelity simulations. Since evaluating a surrogate model is generally much less expensive than executing a high-fidelity simulation, a large number of candidate test scenarios can be assessed before selecting a smaller subset for simulation.

In UAV obstacle-avoidance testing, test scenarios can be constructed by placing obstacles along a flight mission, with the objective of identifying configurations that cause the UAV to approach obstacles closely or exhibit unsafe behaviour.
A flight mission may consist of multiple segments with different positions and orientations, each requiring its own obstacle configurations. Developing an accurate surrogate model for such scenarios requires training data representative of the configurations encountered across these segments.
However, obtaining sufficient training data through high-fidelity simulations can be prohibitively expensive, especially when separate training data must be collected for different flight segments or missions.

An opportunity to address this challenge arises from the geometric symmetries between flight segments. Obstacle configurations associated with different flight segments may have identical relative geometries despite differing in their global positions and orientations.
Such geometrically equivalent configurations can be represented relative to a reference flight segment in a canonical coordinate frame, which we refer to as the canonical flight frame.

In this work, we investigate a surrogate-assisted test-generation approach that exploits these geometric symmetries to reuse a neural-network surrogate across different flight missions.
During training, the surrogate is trained to predict the minimum clearance distance between a UAV and surrounding obstacles for a reference flight mission. This reference flight mission defines the canonical flight frame.
During test generation, candidate obstacle configurations for a target flight mission are transformed into the canonical flight frame and evaluated using the trained surrogate.
Promising configurations are then transformed back into the target mission's coordinate system and evaluated using the high-fidelity simulator.

Geometric equivalence does not necessarily imply equivalent UAV behaviour. The response of an autonomous UAV may also depend on factors such as its incoming velocity, heading, controller state, preceding flight segments, and the surrounding environment. Consequently, the extent to which a surrogate trained in a canonical frame can provide accurate predictions for different missions must be established empirically.

This work investigates the efficiency and effectiveness of canonical
surrogate reuse for simulation-based UAV testing, as well as its
limitations. Specifically, we address the following research questions:
\begin{itemize}
    \item \textbf{RQ1:} How does surrogate prediction accuracy vary across different flight missions?
    \item \textbf{RQ2:} To what extent does an NN-based surrogate improve the efficiency of test generation?
    \item \textbf{RQ3:} How does filtering based on a canonical surrogate affect test-suite diversity?
\end{itemize}

To answer these questions and evaluate our approach, we apply it to efficiently create effective test cases for PX4-Avoidance, a vision-based autonomous obstacle avoidance system developed on top of PX4-Autopilot~\cite{px4avoidance}.
We benchmark our approach against the test case generators that participated in the UAV Testing Competition 2024, 2025 and 2026 versions \cite{khatiri2024sbftuav,khatiri2025sbftuav,khatiri2025icstuav,uysal2026sbftuav,uysal2026icstuav}.

By addressing these research questions, this work makes the following
main contributions: 
\begin{itemize}
    \item A neural network surrogate model that approximates the PX-4 obstacle avoidance system and estimates the minimum clearance distance between a UAV and an obstacle.
    \item An extensive empirical evaluation comparing the efficiency and effectiveness of the surrogate-guided test-case generator with those of state-of-the-art test-case generators.
    \item The limitations of canonical surrogate reuse for simulation-based UAV testing.
    \item A dataset of simulation results for over 2500 different obstacle configurations for a flight mission, which can be used to develop other types of surrogate models.
\end{itemize}

The remainder of this paper is organized as follows. Section~\ref{sec:related-work} reviews related work and Section~\ref{sec:casestudy} introduces the UAV case study. Section~\ref{sec:approach} describes the symmetry-aware testing approach, while Section~\ref{sec:evaluation} reports the empirical evaluation. Section~\ref{sec:discussion} discusses the findings, assumptions, and limitations. Finally, Section~\ref{sec:conclusion} concludes the paper and outlines future work.

\section{Related Work}\label{sec:related-work}

The UAV testing competition has motivated several
strategies for exploring the test-input space. In particular,
DeepHyperion-UAV~\cite{zohdinasab2024deephyperion}
uses illumination search, CAMBA~\cite{deliso2024camba}
employs cost-aware mutation, and
TUMB~\cite{tang2024tumb} uses Monte Carlo Tree Search
to generate obstacle configurations.
These approaches employ different search strategies to identify critical test cases, but evaluating candidate configurations through high-fidelity simulation remains computationally expensive.

To address the cost of simulation-based testing, researchers have explored alternative approaches to test evaluation and generation.
Shrinah and Eder~\cite{shrinah2025pseudorandom} employed a manually constructed low-fidelity path simulator to approximate UAV behaviour at a lower computational cost.

Other approaches have investigated learning-based techniques. Zhu et al.~\cite{zhu2024saflite} proposed SaFliTe, which uses a large
language model to predict test interestingness from textual safety objectives and the current system state.
Peltom\"aki and Porres~\cite{peltomaki2024learning}  proposed WOGAN, which iteratively trains a Wasserstein generative adversarial network to learn a distribution over counterexamples from which failure-inducing inputs can be sampled.

The use of surrogate models to reduce the cost of search-based testing is
well established. Nejati et al.~\cite{nejati2023reflections} provide a
taxonomy of surrogate-assisted search-based testing.
Beglerovic et al.~\cite{beglerovic2017testing}
combined surrogate modelling with stochastic optimisation to search for
fault-revealing regions when testing autonomous vehicles. Similarly,
Haq et al.~\cite{haq2022efficient} employed surrogate-assisted
many-objective optimisation to improve the efficiency of online testing
of DNN-enabled systems. These approaches establish that learned
approximations can reduce the cost of evaluating candidate tests during
search.
In contrast, our work is concerned with the representation used
to train such an approximation: rather than learning separately over
mission-dependent spatial coordinates, geometrically equivalent local
flight configurations are mapped to a common reference frame.

Menghi et al.~\cite{menghi2019approximation} proposed ARIsTEO for testing
compute-intensive cyber-physical models. ARIsTEO constructs a cheaper
surrogate of the model under test, searches this approximation for
failure-revealing inputs, validates candidate failures on the original
model, and refines the approximation when a candidate is spurious. This
approximation--refinement strategy addresses errors in the surrogate
dynamically.

Neural-network-based guidance has also been investigated in evolutionary testing. Zhong et al.~\cite{zhong2021autofuzz} proposed AutoFuzz, which combines
neural-network guidance with evolutionary fuzzing to identify
traffic-violation scenarios for autonomous vehicles.
Similarly, Zohdinasab et al.~\cite{zohdinasab2024focused}  proposed DeepAtash-LR, which integrates a surrogate model into an evolutionary focused-testing process for autonomous driving, directing test generation towards specified combinations of structural and behavioural features. While these approaches demonstrate different ways of using learned models to guide test generation, they do not investigate the reuse of a single surrogate across geometrically related test scenarios through canonical representation.

Geometric symmetry has been widely investigated as a means
of reducing redundant variation in machine learning.
Cohen and Welling~\cite{cohen2016group} introduced
group-equivariant convolutional networks, which incorporate
transformations such as rotations and translations into the
network architecture. Bietti et al.~\cite{bietti2021sample}
studied the sample-complexity benefits of learning under
invariance. These works motivate the exploitation of geometric
symmetry, but do not address surrogate reuse in
simulation-based testing.

Kaba et al.~\cite{kaba2023equivariance} proposed learning
canonicalisation functions that transform inputs into a
canonical representation before prediction. Our approach
follows the same general principle, but determines the
transformation directly from the geometry of a UAV flight
segment rather than learning it. The objective is to reuse
simulation data collected for one reference configuration
when testing different mission segments, while ensuring
that the generated obstacle configurations remain feasible.

Wang et al.~\cite{wang2022approximately} investigated
approximately equivariant networks for dynamical systems
in which geometric symmetry is imperfect. This limitation
is also relevant to our work: geometrically equivalent
obstacle configurations may produce different UAV
behaviour because of controller state, preceding flight
segments, or environmental constraints. We therefore
evaluate cross-mission surrogate accuracy empirically
rather than assuming that geometric equivalence guarantees
identical simulation outcomes.

Existing research has established surrogate-assisted testing
as a means of reducing expensive simulations \cite{nejati2023reflections,beglerovic2017testing,
      menghi2019approximation,zohdinasab2024focused} and
canonicalisation as a means of exploiting geometric
symmetry \cite{cohen2016group,kaba2023equivariance,bietti2021sample}. 
Our work combines these established ideas
to investigate whether a safety surrogate trained on a
single canonical UAV flight configuration can be reused
across different mission segments without retraining.

\section{Case Study}\label{sec:casestudy}
We use the UAV Testing Competition as a case study for
evaluating our symmetry-aware surrogate-assisted testing
approach. The competition was introduced at SBFT in 2024 and subsequently organized
at both SBFT and ICST in 2025 and 2026.
The competition provides a benchmark for comparing automated test-case generation techniques for autonomous UAVs.

The system under test is PX4-Avoidance, an obstacle-avoidance system built on PX4-Autopilot.
Test cases specify the geometry and placement of cuboid obstacles for a given flight mission.
According to the competition constraints, valid test cases must have non-overlapping obstacles that lie entirely within the designated obstacle-placement area, defined as a $70,\mathrm{m} \times 40,\mathrm{m}$ rectangle.

Test cases are used to expose unsafe behaviours by executing the given flight mission in the presence of the obstacles defined by the test case using Aerialist, a simulation-based UAV testing framework for PX4-Autopilot \cite{khatiri2024aerialist}.

The competition scores test generators based on the criticality and diversity of their test suites.
Criticality is measured by the minimum clearance distance between the UAV and the obstacles, while diversity is assessed by comparing test-case similarities within the suite.
We use the competition missions and evaluation criteria to assess the efficiency and effectiveness of our approach.

\section{Symmetry-Aware Surrogate-Assisted Testing}\label{sec:approach}
\subsection{Overview}
Our approach generates a set of tests that are predicted to cause a UAV to fly close to an obstacle for a given flight mission.  
Each test case consists of two obstacles, each parameterised by five values: width, height, centre coordinates, and rotation.  Together, the two obstacles form a ten-element test vector.  

The approach uses a neural network surrogate model to estimate the minimum clearance distance between a UAV and an obstacle during a mission.  The model is trained using randomly generated tests from a single canonical mission, which serves as the reference mission during training and inference.

A transformation is found between any arbitrary mission and the canonical mission.  This transformation enables tests to be generated within the surrogate model’s training domain, where prediction accuracy is expected to be highest.  

During test generation, batches of tests are first produced using a constrained-random generator.  The surrogate model is then used in a filtering mechanism to select tests that are both critical and diverse.  Critical tests are defined as those for which the minimum clearance distance is below 1.5m.  Diverse tests are defined as those that increase the structural variability of non-overlapping geometries in the test suite's input space (Equation~\ref{equ:input_diversity}).
Finally, the selected test cases are executed in the high-fidelity
simulator to obtain their actual minimum clearance distances and verify
the surrogate's predictions.

\subsection{Canonical Representation of Flight Segments}
\begin{figure}[h]
    
        \includegraphics[trim={0.0cm, 0.0cm, 0.0cm, 0.0cm}, clip, width=\linewidth]{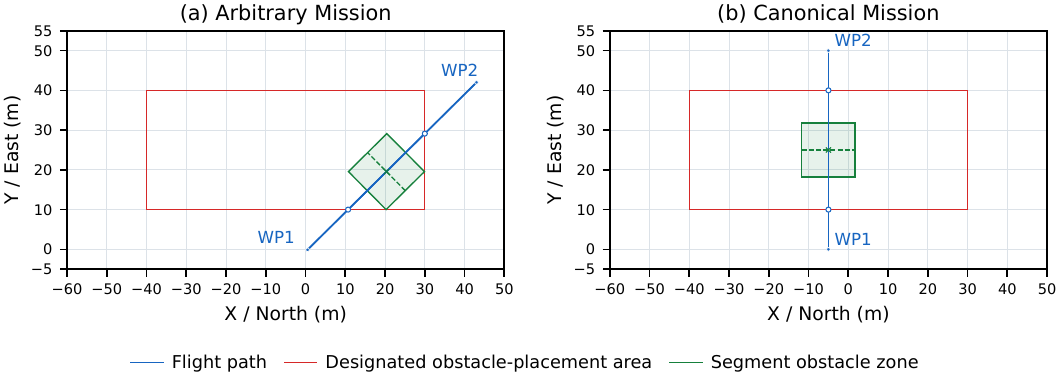}
   
    \caption{A segment of an arbitrary mission path is transformed to the canonical mission representation.  The segment-placement zone is chosen in the arbitrary mission frame such that when transformed between representations, it always remains inside the designated obstacle-placement area.}
    \label{fig:mission_transformation}

\end{figure}

To ensure the surrogate model is always used within its training domain, each arbitrary mission is transformed into a canonical mission. 
The canonical mission is a single vertical flight path, centred horizontally within the designated obstacle-placement area, as illustrated in Figure~\ref{fig:mission_transformation}.

\clearpage
\noindent The transformation consists of three steps:
\begin{enumerate}
    \item For each flying segment between two consecutive waypoints, determine a segment-placement zone by constructing the largest rectangle, symmetric about the segment, that lies entirely within the flight-testing arena.
    \item  Select the mission segment whose segment-placement zone has the greatest area.
    \item Apply a two-dimensional translation and rotation to map both the mission segment and its associated segment-placement zone into the canonical representation.
\end{enumerate}
The transformed segment-placement zone defines the geometric constraints used during random obstacle generation.  Depending on the mission geometry, this zone may occupy only a subregion of the obstacle placement zone used to train the surrogate model. Consequently, all generated obstacles remain within the surrogate model's training domain.

After test generation, the inverse transformation is applied, mapping the generated obstacles from the canonical representation back to the original mission frame.  Tests expressed in the original mission frame are then evaluated using the Aerialist simulator.

\subsection{Surrogate Model Training}
\begin{figure*}[t]
    \centering
    \includegraphics[trim={0.0cm, 1.5cm, 0.0cm, 0.0cm}, clip, width=\textwidth]{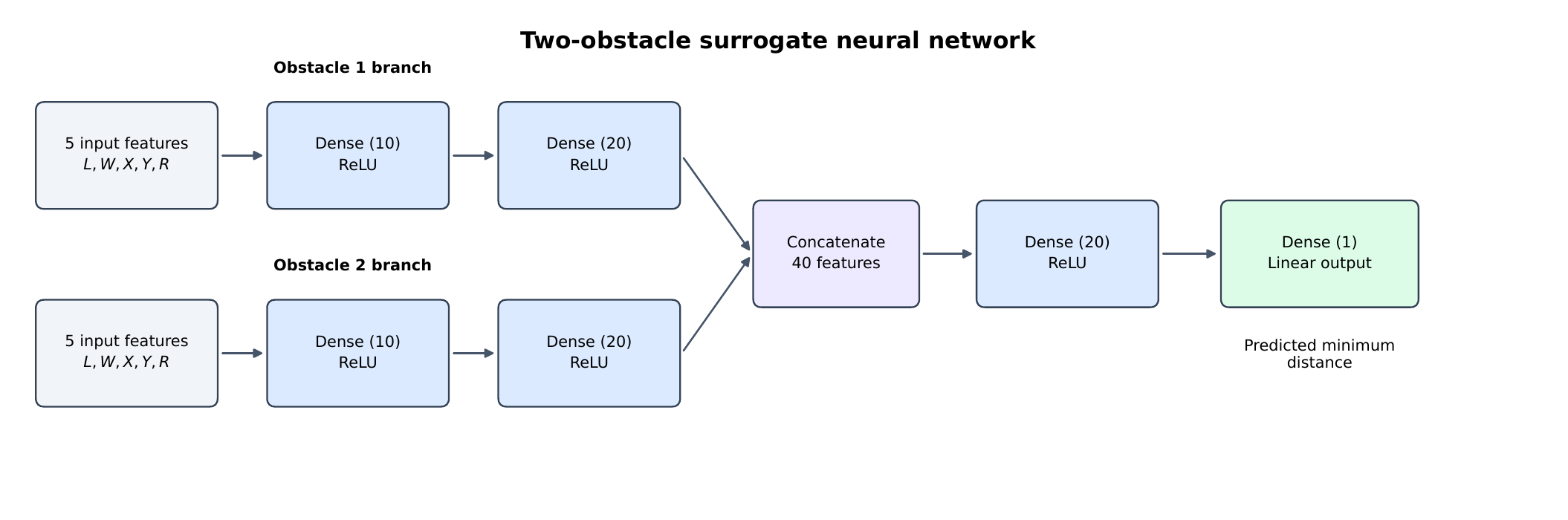}
    \caption{The architecture of the surrogate model.}
    \label{fig:nnarchitecture}
    
\end{figure*}

The surrogate model is a neural-network regression model that predicts the minimum clearance distance between the UAV and any obstacle during the mission.  The model takes the geometry of two obstacles as input and outputs an estimated minimum clearance.  
Its architecture, depicted in Figure~\ref{fig:nnarchitecture}, comprises two parallel subnetworks, one for each obstacle, with sizes of 5 input, 10 hidden and 20 output layer neurons.
    
The outputs of the subnetworks are concatenated and passed to a fully connected network with 40 input, 20 hidden and 1 output neuron.  
ReLU activations are used for all hidden layers, while the output layer uses a linear activation.  This architecture was selected empirically using validation performance.

The surrogate model’s dataset is created from 2716 random test cases of two obstacles within the designated obstacle-placement area.
Eighty percent of the random test cases were used for training, and the remaining twenty percent were used for validation.

The dataset was sampled from the space of all possible obstacle configurations by a specialist constrained-random generator that employs semantic sampling rather than direct geometric sampling.  Scenario features such as obstacle intrusion into the flight path, size, aspect ratio, rotation, relative neighbour direction, and passage width were discretised into a set of categories. 
For example, obstacle rotation was classified into low ($0^\circ$ to $45^\circ$) and high ($45^\circ$ to $90^\circ$) categories, while relative obstacle placement was classified into one of eight neighbouring sectors. 
Data generation first uniformly sampled a combination of these categories, one from each feature, and subsequently randomly sampled geometric values from the corresponding parameter ranges. 
This approach ensured balanced coverage of distinct scenario types and avoided over-representation of geometrically similar configurations.

This sampling process was observed to produce highly skewed distributions for the obstacle widths and the $x$ coordinate of obstacle one.   Hence, to reduce the effect of skewness, a log transformation was applied to these features prior to standardisation.  All ten geometry features were then normalised using feature-wise mean and standard deviation statistics calculated from the training set.
Each obstacle configuration was evaluated using Aerialist to obtain the ground-truth minimum clearance distance used in training.

The network was trained for 200 epochs using the Adam optimiser, a batch size of 128, and mean absolute error (MAE) as the loss function.

\subsection{Feasibility-Constrained Obstacle Generation}
The trained surrogate model is used with a constrained-random generator that generates obstacle geometries (tests).  The constraints for this random generator are derived from the segment-placement zone for each mission, transformed into the canonical representation.  These constraints are applied to the generation of the ten obstacle parameters to ensure that each obstacle pair do not overlap and remain entirely within the placement zone.

\subsection{Surrogate-Guided Test Selection}
Test generation is split into two phases.  In both phases, a fixed-size leaderboard is maintained to store the most critical and diverse tests discovered so far.  

\textbf{Phase one} focuses on identifying tests with low predicted obstacle clearances.  A batch of tests is repeatedly generated by the constrained-random generator.  For each test, the surrogate model predicts the minimum obstacle clearance distance.  The leaderboard is then updated to retain the tests with the smallest predicted clearances seen so far, while tests with larger predicted clearances are discarded.  This process of generation and selection is repeated 100 times.  At the end of phase one, the leaderboard contains tests with small clearances, but potentially poor diversity.

\textbf{Phase two} shifts the emphasis to improving leaderboard diversity while retaining only critical tests.  The constrained-random generator generates new batches of tests to improve test suite diversity, and the surrogate model filters out any non-critical tests (predicted minimum clearance distance $>$ 1.5m), leaving a set of candidate tests.  For each candidate set, a series of alternative leaderboards are created by replacing one existing test with a candidate test, and evaluating the diversity of the resulting leaderboard using Equation~\ref{equ:input_diversity}, described in Section~\ref{sec:evaluation}.   The leaderboard with the highest diversity is retained for the next iteration.  The process of candidate generation, selection and diversity evaluation is repeated for 100 rounds, producing a leaderboard that balances both test criticality and diversity.

\textbf{Phase three} verifies the predicted criticality of test cases in the leaderboard.
Each test in the final leaderboard is evaluated using the high-fidelity simulator to obtain a more accurate estimate of minimum clearance.  
Our approach balances computational efficiency and prediction accuracy.  In the first two phases, the approach rapidly explores a large space of potential tests, while the computationally expensive simulator is reserved for the relatively small set of tests selected for the final leaderboard in phase three.
Consequently, simulation effort is concentrated on the most promising tests, improving confidence in the final ranking without requiring exhaustive simulation of all considered tests.

\section{Empirical Evaluation}\label{sec:evaluation}
\subsection{Experimental Setup}

The performance of our approach was benchmarked against submissions from the 2024, 2025 and 2026 editions of the ICST/SBFT UAV testing competitions. 
The benchmark comprises the flight mission set, simulation budget and scoring criteria.
We aligned our evaluation with these components to enable a direct comparison between our proposed approach and the prior competition entrants.
We also recalculated the scores of the benchmarks using the publicly available simulation artifacts and scoring criteria. 
All experiments were conducted on a machine equipped with an Intel Core i7-10870H processor (8 cores, 16 threads) and 32 GB of RAM, running Ubuntu 20.04.6 LTS. UAV simulations were executed in Docker containers using Aerialist version 0.2.02 (Git revision \texttt{f44ea6a}) with Ubuntu 18.04.5 LTS as the container operating system.
The datasets, source code, and experimental artifacts supporting this study are available on Zenodo~\cite{Shrinah2026UAVArtifacts}.

Each component of the benchmark is summarised in this section, with complete details of the format and scoring for each competition year found in \cite{khatiri2024sbftuav,khatiri2025sbftuav,khatiri2025icstuav,uysal2026sbftuav,uysal2026icstuav}.

\subsubsection{flight mission}

\begin{figure*}[t]
    \centering
    \includegraphics[width=\textwidth]{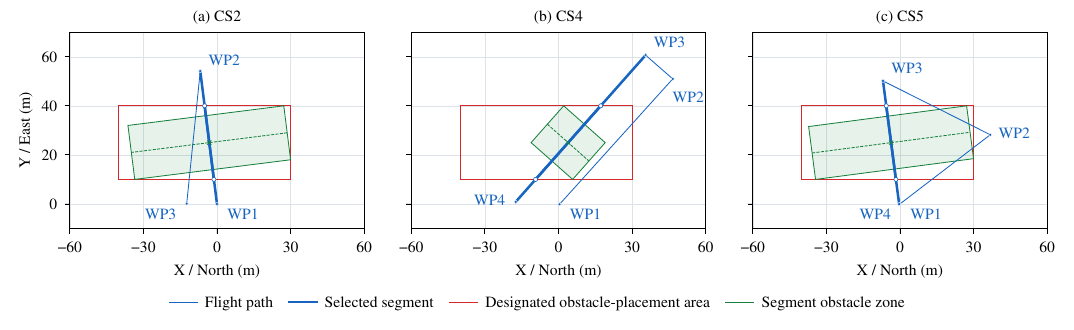}
   
    \caption{The flight missions, path segments, and segment-placement zones for transformation to the canonical mission.}
    \label{fig:missions}
\end{figure*}

The three flight missions common to all competitions, namely CS2, CS4 and CS5, were selected for evaluating our approach.  Each mission varies in the number and layout of waypoints, and the extent to which the flight-path segments overlap the valid obstacle-placement area.  See Figure~\ref{fig:missions}.

\subsubsection{Simulation Budget}~\label{sec:simulation_budget}
The simulation budget defines the maximum number of test cases that can be evaluated using Aerialist during test generation, prior to submission and scoring. In accordance with the competition requirements, we use simulation budgets of 200 for 2024 and 100 for 2025 and 2026. Consequently, we execute our approach twice: once with a budget of 200 to generate a test suite for comparison against the 2024 entrants, and once with a budget of 100 for comparison against the 2025 and 2026 entrants. In both runs, the leaderboard size is set equal to the corresponding simulation budget.

\subsubsection{Scoring Criteria}
The scoring is based on simulations of the submitted test suite using Aerialist.
Each test case is simulated three times, and averages are calculated for the minimum clearance distance, flight path, and execution time.  
Each simulation run was limited to 500 seconds. Runs that exceeded this limit before completing the required flight mission were terminated and recorded as timed out.
Different scoring criteria is used for each year, reflecting changes made by the competition organisers intended to improve the assessment criteria.  

\textbf{2024:} The submitted test suite ($S$) is split into two groups.  One group, $S_s$ is scored directly, and the other group, $S_r$ is scored based on an average score from the first group.  The competition organisers chose this approach to reduce the computational time needed to score an entire test suite.

The final score integrates the test scores for simulated and unsimulated tests and a test similarity penalty:

\begin{equation}
\begin{split}
        suite\_score(s) = (&test\_score(S_s) + rest\_score(S_r))\\
                            &\times similarity\_penalty(S_s)
\end{split}
\end{equation}

For the first group, $S_s$, 20 tests are selected at random from the submitted test suite.
Each of these test cases was checked for validity. Obstacles must not overlap or extend outside the valid obstacle-placement area. 
Valid test cases are then simulated three times with Aerialist, and the average path, execution time and minimum obstacle clearance distance are recorded.  

Runs of the simulated tests were awarded 5 points if the minimum clearance distance is less than 0.25m, 2 points if the distance is between 0.25 and 1.00m, 1 point if the distance is between 1.00m and 1.5m and zero otherwise.  Tests with minimum clearance distances less than 1.5m are denoted critical tests.
The points from the three simulations for each test case are averaged.  

The $test\_score$ for each test case is then calculated using its average point ($avg\_point$), the number of obstacles used ($obst$) and the average mission flight time ($avg\_time$) in minutes.
\begin{equation}
    test\_score(t) = \frac{10\times avg\_point(t)}{\#obst(t)^2 \times avg\_time(t)}     
\end{equation}
The test score of the set of selected tests $test\_score(S_s)$ is the sum of the test scores of the individual tests $test\_score(t)$.
The remaining unsimulated tests are scored based on an average from the simulated set.
\begin{equation}
    rest\_score(S_r) = |S_r| \times \frac{\sum_{t}test\_score(t)}{2|S_s|}
\end{equation}
If less than 20 tests are submitted, then  $|S_r|=0$ and no score is given for unsimulated tests.

Test diversity is calculated using the Dynamic Time Warping distance between the trajectories of any two simulated tests.  A test is counted as similar if the DTW distance is less than 75.  This value was chosen for compatibility with the 2024 competition artifacts.
The similarity penalty is calculated as the complement of the ratio of similar test cases from the total number of simulated test cases:
\begin{equation}
    similarity\_penalty(S_s) = 1 - \frac{\#similar\_tests(S_s)}{|S_s|}
\end{equation}

\textbf{2025:} The scoring method for 2025 was updated as follows. The set of 20 tests for simulation ($S_s$) were not selected randomly. Instead, the tests for simulation are determined by the entrant’s test generator.
For our approach, these are the 20 tests with the maximum $test\_score(t)$ (The formula for calculating  $test\_score(t)$ did not change in 2025.), where the test score is calculated using the results of the simulations performed during phase three using the simulation budget. 
Furthermore, tests not selected for simulation are not awarded any score.

In contrast to 2024, the 2025 competition did not assign an overall score to each test suite. Instead, participants were ranked separately for each mission based on their test suite score, which is called failure score in 2025, and diversity scores. The ranks for each category were summed across missions to determine the category standings.
The overall ranking gave failure detection (test criticality) performance three times the weight of test diversity:
\begin{equation}
\begin{split}
        overall\_rank(S_s) = &0.75 \times failure\_rank(S_s) \\  + &0.25 \times diversity\_rank(S_s)
\end{split}
\end{equation}

\noindent Diversity is scored as:
\begin{equation}
    diversity\_score(S_s) = 1 - \frac{1}{|S_s|}\sum_{t_i, t_j\in S_s}similarity(t_i, t_j)
\end{equation}
where:
\begin{equation}
    similarity(t_i, t_j) = \frac{|area(t_i)\cap area(t_j)|}{|area(t_i)\cup area(t_j)|}
\end{equation}

\text{\textbf{2026:}} 
The diversity score is split into two components: input diversity, which measures variation in test geometries, and output diversity, which measures variation in the resulting UAV flight paths.
The overall ranks of the test suites are calculated as follows:
\begin{equation}
\begin{split}
        overall\_rank(S_s) = &0.50 \times failure\_rank(S_s) \\  + &0.25 \times input\_diversity\_rank(S_s) \\  + &0.25 \times output\_diversity\_rank(S_s)
\end{split}
\end{equation}

The selection of the simulation set, $S_s$, the calculation of $test\_score$, and ranking approach follow the same method used in 2025.  Input diversity is calculated as:
\begin{equation}~\label{equ:input_diversity}
    input\_diversity = D_{union}(1+CV)
\end{equation}
Where:
\begin{equation}
    D_{union} = \frac{A(\cup_{i=1}^{n}Gi)}{1/n\sum_{i=1}^{n}A(G_i)}
\end{equation}
and the structural variability, $CV$, is defined as:
\begin{equation}
    CV = \frac{\sqrt{\frac{1}{n-1}\sum_{i=1}^{n}(A(G_i)-\bar{A})^2}}{\bar{A}}, where \bar{A}=\frac{1}{n}\sum_{i=1}^{n}A(G_i)
    \end{equation}
and $A(G_i)$ is the area of a single obstacle and $n$ is the number of obstacles in $S_s$.  

The output diversity of a test suite is calculated from the UAV's Aerialist simulated flight trajectories in the presence of obstacle placements.

For each test, $j$, and its average simulated flight trajectory, $\bar{\tau_j}$, the dynamic time warping (DTW) distance, $\delta_j$ is calculated as the minimum pairwise distance to any other test's average trajectory $\bar{\tau_k}$:
\begin{equation}
    \delta_j = \min_{k=1, k\neq j}^{m}DTW(\bar{\tau_j}, \bar{\tau_k})
\end{equation}

Output diversity is then defined as the minimum DTW distance, averaged across all tests in the submitted test suite. 
\begin{equation}
    ouput\_diversity(S_s) = \frac{1}{m}\sum_{j=1}^{m}\delta_j
\end{equation}

\section{Results and Discussion}\label{sec:discussion}

The proposed approach demonstrated strong performance across the three competition editions as shown in Table~\ref{tab:benchmark-2024}, Table~\ref{tab:benchmark-2025}, and Table~\ref{tab:benchmark-2026}. Under the 2024 evaluation criteria, it achieved the highest score for all three missions. In 2025, it obtained the best overall ranking, achieving the highest failure scores for CS2 and CS5 and the second-highest for CS4. In 2026, the approach ranked second overall. 

A noteworthy observation is that our approach achieved the highest precision in all three benchmark evaluations.
We define precision as the proportion of test cases identified as critical by a generator that are subsequently confirmed as critical by the high-fidelity simulator.
To account for non-determinism in the simulation environment, criticality is confirmed by taking the average minimum clearance distance from three high-fidelity simulations.  

In 2024, our approach submitted 228 test cases across the three missions. Of the 60 cases selected for evaluation, 53 were confirmed critical, yielding a precision of 88.3\%. This exceeded the 79.2\% precision achieved by the next-best competitor in terms of precision, TAIist, which submitted 24 test cases, 19 of which were confirmed critical.
In 2025, our approach submitted 146 test cases and achieved a precision of 96.7\% (58 out of 60 evaluated cases), compared with 66.7\% (40 out of 60) for PALM, which submitted 216 test cases. In 2026, the same 146 test cases and their evaluation from 2025 were reused for benchmarking the precision of our approach. 
Thus, the 96.7\% (58/60) precision from the total of 146 submitted test cases exceeded the 76.8\% 
precision (43 out of 56 evaluated cases) achieved by Intelligent-UAV-Gen, which submitted 159 test cases.

Overall, the results show that our approach consistently achieved higher precision than the strongest competing generators across all three benchmarks. Its high precision, together with the large number of generated test cases, indicates that our approach has high repeatability.

Another particularly interesting observation concerns the
relationship between input and output diversity.
In 2026, our approach achieved the highest input-diversity score for all three missions. This is consistent with the design of the second test-generation phase, which explicitly maximises input diversity.
With regard to output diversity, our approach ranked third for CS2 and fourth for CS4 and CS5.
The relatively low output diversity of our approach is explained in the discussion of RQ3 in Section~\ref{RQ3}.
These contrasting results indicate that generating geometrically diverse obstacle configurations does not necessarily produce diverse UAV flight trajectories.

Another observation is that output diversity varies across missions. In 2026, three of the four approaches achieved their lowest output-diversity scores on CS4.
One potential explanation is that among the three missions, CS4 has the largest proportion of its planned flight path outside the obstacle placement zone (56.84\%, compared with 44.51\% for CS2 and 51.84\% for CS5). Consequently, a smaller proportion of its route lies within the region where obstacles can be placed, potentially limiting opportunities for obstacle-induced behavioural variation.

\subsection{RQ1: How does the Surrogate Prediction Accuracy vary for different flight missions?}
Prediction accuracy is quantified using the root mean squared error (RMSE) between the NN surrogate's predicted minimum clearance distance and the ground-truth clearance obtained from Aerialist.  The assessment is made using 300 tests, which are the combined set of tests generated by the proposed approach for the 2024 and 2025/2026 benchmarks.  None of these tests were included in either the training or validation sets, ensuring the evaluation measures the model's performance on previously unseen test cases. 

To provide a reference point, the trained surrogate model was also evaluated on its validation dataset, which was generated by the constrained-random generator for the canonical mission.  Comparing these results enables the accuracy of the surrogate model on generated tests to be contrasted with its accuracy on tests drawn from the same distribution as the training process.

\begin{table}[h]
    \centering
    \begin{tabular}{c|c}
         Mission & RMSE \\
         \hline
         CS2  &  1.35m \\
         CS4 &  0.71m \\
         CS5 &  1.46m \\
         Validation Set & 0.50m \\
    \end{tabular}
    \caption{Surrogate Model Prediction Error}
    \label{tab:prediction_rmse}
\end{table}

Since missions two, four and five are transformed into the canonical representation prior to prediction, it might be expected that their RMSE values would be similar to that observed for the validation set.  However, the RMSE is higher for all three missions than for the validation set. This difference can be explained by how the ground truth clearances are obtained.

The validation data were generated for the canonical mission, which contains only a single flight-path segment.
However, the benchmark missions may feature multiple flight path segments that pass through the obstacle placement zone.  These additional segments can interact with the obstacles and influence the minimum observed clearance during simulation.  Such interactions are not represented in the surrogate model's training data and therefore are a source of error for its predictions.

The characteristics of the three benchmark missions support this explanation.  Of the three missions, only mission four has a single flight path segment through the obstacle placement zone (See Figure~\ref{fig:missions}-b).  In contrast, mission two crosses its respective obstacle placement zone twice and mission five passes through it three times.  As the number of flight-path crossings through the obstacle placement zone increases, so does the opportunity for obstacle interactions that are not represented in the canonical mission training data.

\subsection{RQ2: Test-Generation Efficiency}

The efficiency of a test generation method can be measured by how many critical test cases are generated from a given testing budget. We call this ratio a hit rate.

We compare the hit rate of our approach with the constrained-random generator used to create the surrogate model's training and validation datasets. 
During dataset generation, the constrained-random generator produced 2716 test cases for the canonical flight mission of which 265 are critical,~i.e. it has approximately a 10\% hit rate.
This dataset does not exclude tests that cause the UAV to exceed the allowed time limit of 500 seconds as specified in the competition scoring criteria.  Consequently, the constrained-random generator's hit rate is an upper limit.

In contrast, the surrogate-assisted pseudo-random generator used a total simulation budget of 900 to identify 375 critical test cases across the benchmark missions, corresponding to a hit rate of approximately 42\%.
The overall hit rate of our approach is calculated as the number of critical test cases identified by our approach divided by the sum of the simulation budgets for all missions for 2024 and 2025/26 benchmarks.
The surrogate-assisted approach achieved a critical-test-case hit rate approximately four times that of the constrained-random generator.

It is also noticeable from Table~\ref{tab:test-case-outcomes} that the hit rate for our approach varied by mission.
Taking the average hit rate across the 300 simulations for each mission (200 from 2024 and 100 from 2025/26), produced hit rates of 34\%, 49\% and 42\% for missions CS2, CS4 and CS5, respectively.  Each of these average hit rates is higher than the hit rate achieved by the constrained-random generator.

Note that we compared the efficiency of our approach against a constrained-random generator rather than a purely random generator, as the latter would have produced far fewer critical test cases.
The constrained-random generator is more efficient than a purely random generator that does not embed domain knowledge into the generation process.

\begin{table*}[t]
\centering
\caption{Test case outcomes and hit rates across benchmark years and flight missions.}
\label{tab:test-case-outcomes}
\small
\setlength{\tabcolsep}{5pt}
\begin{tabular}{@{}c c c c c c c@{}}
\toprule
\textbf{Benchmark} &
\textbf{Simulation} &
\textbf{Flying} &
\multicolumn{3}{c}{\textbf{Number of Test Cases}} &
\textbf{Hit Rate} \\
\textbf{Year} &
\textbf{Budget} &
\textbf{Mission} &
\textbf{Timed Out} &
\textbf{Non-Critical} &
\textbf{Critical} &
\textbf{(\%)} \\
\midrule
\multirow{3}{*}{2024}
& 200 & CS2 & 22 & 116 & 62 & 31 \\
& 200 & CS4 & 1 & 111 & 88 & 44 \\
& 200 & CS5 & 15 & 106 & 79 & 40 \\
\midrule
\multirow{3}{*}{2025/2026}
& 100 & CS2 & 17 & 43 & 40 & 40 \\
& 100 & CS4 & 5 & 36 & 59 & 59 \\
& 100 & CS5 & 13 & 40 & 47 & 47 \\
\bottomrule
\end{tabular}
\end{table*}

\subsection{RQ3: How Does Canonical Surrogate-Based Filtering Affect Test-Suite Diversity?}\label{RQ3}

This research question investigates whether surrogate-based filtering restricts the behavioural diversity of resulting test suites, given that the model is trained on pseudo-randomly generated test cases for a single canonical flight mission.

As discussed in RQ1, the surrogate may produce more accurate predictions for flight missions whose characteristics resemble those of the canonical mission used during training. However, differences in prediction accuracy across missions do not necessarily imply that the filtering process systematically favours particular types of test cases. 
Prediction accuracy and selection bias are distinct concerns.
Nevertheless, our approach achieved one of the lowest behavioural diversity scores among the participants in the 2026 competition.

One possible explanation for the low behavioural diversity is the restricted range of obstacle configurations produced by the constrained-random generator in the first phase of our approach. To remain within the surrogate's training domain, obstacle configurations are restricted to a predefined region relative to the flight path. These constraints limit the variety of configurations that can be generated and may reduce interactions with other flight segments, potentially resulting in less diverse UAV trajectories.

Another possible explanation concerns obstacle size. Our approach generated comparatively small obstacles, which may cause relatively minor deviations from the intended flight path and reduce the likelihood of interactions with other flight segments.

Overall, the relatively low behavioural diversity achieved by our approach may be associated more strongly with the restrictions imposed during initial test-case generation than with surrogate-based filtering.
However, confirming the effect of these restrictions on diversity would require comparing the output diversity of two test suites, one generated by a purely random test generator and the other using the constrained-random generator under otherwise identical conditions.

\begin{table*}[t]
\centering
\caption{Benchmark results for the 2024 evaluation. Each entry reports the number of test cases (critical test cases) or the score (score rank).}
\label{tab:benchmark-2024}
\small
\setlength{\tabcolsep}{5pt}
\resizebox{\textwidth}{!}{%
\begin{tabular}{@{}l *{7}{c}@{}}
\toprule
& \multicolumn{2}{c}{CS2} & \multicolumn{2}{c}{CS4} & \multicolumn{2}{c}{CS5} & \multicolumn{1}{c}{Overall} \\
\cmidrule(lr){2-3}\cmidrule(lr){4-5}\cmidrule(lr){6-7}\cmidrule(l){8-8}
\textbf{Tool} & \makecell{\textbf{Tests}\\\textbf{(critical)}} & \makecell{\textbf{Score}\\\textbf{(rank)}}
& \makecell{\textbf{Tests}\\\textbf{(critical)}} & \makecell{\textbf{Score}\\\textbf{(rank)}}
& \makecell{\textbf{Tests}\\\textbf{(critical)}} & \makecell{\textbf{Score}\\\textbf{(rank)}}
& \makecell{\textbf{Score sum}\\\textbf{(rank sum)}} \\
\midrule
Our approach & 62 (19) & 18.27 (1) & 88 (18) & 20.32 (1) & 78 (16) & 19.21 (1) & 57.80 (3) \\
WOGAN-UAV & 61 (14) & 12.55 (2) & 81 (4) & 2.35 (4) & 39 (12) & 8.40 (2) & 23.30 (8) \\
CAMBA & 36 (7) & 11.84 (3) & 33 (0) & 0.00 (7) & 11 (1) & 3.16 (3) & 15.00 (13) \\
TUMB & 69 (2) & 0.12 (8) & 135 (16) & 7.12 (3) & 114 (7) & 2.73 (4) & 9.97 (15) \\
DeepHyperion & 2 (2) & 1.22 (6) & 10 (8) & 7.74 (2) & 22 (4) & 1.08 (7) & 10.04 (15) \\
AmbieGen & 30 (12) & 2.82 (5) & 36 (4) & 0.86 (5) & 65 (5) & 1.22 (6) & 4.90 (16) \\
TAIist & 7 (7) & 0.81 (7) & 6 (4) & 0.43 (6) & 11 (8) & 1.57 (5) & 2.81 (18) \\
Surrealist & 10 (?) & 5.03 (4) & 10 (?) & 0.00 (7) & 1 (?) & 0.75 (8) & 5.78 (19) \\
\bottomrule
\end{tabular}
}%
\vspace{2pt}
\parbox{\textwidth}{\footnotesize\emph{Note:} ``?'' indicates that the number of critical test cases was not available in the 2024 competition artifacts.}
\end{table*}
  \begin{table*}[t]
\centering
\caption{Benchmark results for the 2025 evaluation. Each entry reports test cases (critical test cases) or a failure/diversity score (rank).}
\label{tab:benchmark-2025}
\small
\setlength{\tabcolsep}{3pt}
\resizebox{\textwidth}{!}{%
\begin{tabular}{@{}l *{10}{c}@{}}
\toprule
& \multicolumn{3}{c}{CS2} & \multicolumn{3}{c}{CS4} & \multicolumn{3}{c}{CS5} & \multicolumn{1}{c}{Overall} \\
\cmidrule(lr){2-4}\cmidrule(lr){5-7}\cmidrule(lr){8-10}\cmidrule(l){11-11}
\textbf{Tool} & \makecell{\textbf{Tests}\\\textbf{(critical)}} & \makecell{\textbf{Failure}\\\textbf{(rank)}} & \makecell{\textbf{Diversity}\\\textbf{(rank)}} & \makecell{\textbf{Tests}\\\textbf{(critical)}} & \makecell{\textbf{Failure}\\\textbf{(rank)}} & \makecell{\textbf{Diversity}\\\textbf{(rank)}} & \makecell{\textbf{Tests}\\\textbf{(critical)}} & \makecell{\textbf{Failure}\\\textbf{(rank)}} & \makecell{\textbf{Diversity}\\\textbf{(rank)}} & \makecell{\textbf{Rank}\\\textbf{value}} \\
\midrule
Our approach & 40 (20) & 11.90 (1) & 0.94 (2) & 59 (19) & 9.14 (2) & 0.85 (4) & 47 (19) & 10.89 (1) & 0.93 (1) & 4.75 \\
Pseudo-Random & 31 (7) & 6.54 (3) & 0.86 (5) & 66 (12) & 9.30 (1) & 0.87 (3) & 17 (5) & 2.71 (3) & 0.82 (5) & 8.5 \\
Evolv-1 & 57 (12) & 9.02 (2) & 0.88 (4) & 22 (2) & 1.08 (6) & 0.82 (5) & 37 (6) & 2.93 (2) & 0.85 (4) & 10.75 \\
TGen-UQ & 5 (1) & 0.50 (5) & 1.00 (1) & 6 (1) & 1.44 (5) & 0.94 (1) & 6 (0) & 0.00 (6) & 0.89 (2) & 13.00 \\
PALM & 58 (1) & 0.14 (7) & 0.57 (6) & 82 (20) & 3.58 (3) & 0.42 (6) & 76 (19) & 0.80 (4) & 0.46 (6) & 15.00 \\
Surrealist & 52 (11) & 4.16 (4) & 0.21 (7) & 9 (3) & 1.93 (4) & 0.20 (7) & 9 (1) & 0.11 (5) & 0.18 (7) & 15.00 \\
OptObstacles & 3 (1) & 0.44 (6) & 0.93 (3) & 7 (0) & 0.00 (7) & 0.92 (2) & 2 (0) & 0.00 (6) & 0.89 (2) & 16.00 \\
\bottomrule
\end{tabular}%
}
\vspace{2pt}
\end{table*}

   \begin{table*}[t]
\centering
\caption{Benchmark results for the 2026 evaluation. Each entry reports test cases (critical test cases) or a failure/diversity score (rank).}
\label{tab:benchmark-2026}
\small
\setlength{\tabcolsep}{2.5pt}
\resizebox{\textwidth}{!}{%
\begin{tabular}{@{}l *{13}{c}@{}}
\toprule
& \multicolumn{4}{c}{CS2} & \multicolumn{4}{c}{CS4} & \multicolumn{4}{c}{CS5} & \multicolumn{1}{c}{Overall} \\
\cmidrule(lr){2-5}\cmidrule(lr){6-9}\cmidrule(lr){10-13}\cmidrule(l){14-14}
\textbf{Tool} & \makecell{\textbf{Tests}\\\textbf{(critical)}} & \makecell{\textbf{Failure}\\\textbf{(rank)}} & \makecell{$\boldsymbol{D}_{\mathrm{in}}$\\\textbf{(rank)}} & \makecell{$\boldsymbol{D}_{\mathrm{out}}$\\\textbf{(rank)}} & \makecell{\textbf{Tests}\\\textbf{(critical)}} & \makecell{\textbf{Failure}\\\textbf{(rank)}} & \makecell{$\boldsymbol{D}_{\mathrm{in}}$\\\textbf{(rank)}} & \makecell{$\boldsymbol{D}_{\mathrm{out}}$\\\textbf{(rank)}} & \makecell{\textbf{Tests}\\\textbf{(critical)}} & \makecell{\textbf{Failure}\\\textbf{(rank)}} & \makecell{$\boldsymbol{D}_{\mathrm{in}}$\\\textbf{(rank)}} & \makecell{$\boldsymbol{D}_{\mathrm{out}}$\\\textbf{(rank)}} & \makecell{\textbf{Rank}\\\textbf{value}} \\
\midrule
Intelligent-UAV-Gen & 66 (20) & 19.23 (1) & 10.74 (2) & 4.13 (2) & 16 (4) & 3.12 (2) & 6.89 (3) & 3.07 (2) & 77 (19) & 25.33 (1) & 8.64 (2) & 3.11 (3) & 5.50 \\
Our approach & 40 (20) & 11.90 (3) & 39.11 (1) & 2.72 (3) & 59 (19) & 9.14 (1) & 16.85 (1) & 1.37 (4) & 47 (19) & 10.89 (2) & 32.76 (1) & 2.41 (4) & 6.50 \\
TLQD & 100 (7) & 6.66 (4) & 9.79 (3) & 23.31 (1) & 100 (4) & 1.11 (4) & 16.74 (2) & 14.44 (1) & 100 (10) & 9.13 (3) & 4.74 (3) & 39.62 (1) & 8.25 \\
Surrealist & 80 (17) & 18.25 (2) & 2.80 (4) & 1.00 (4) & 13 (4) & 2.44 (3) & 2.68 (4) & 2.67 (3) & 79 (0) & 0.00 (4) & 3.00 (4) & 4.26 (2) & 9.75 \\
\bottomrule
\end{tabular}%
}
\vspace{2pt}
\end{table*}

\section{Conclusion and Future Work}\label{sec:conclusion}
This paper investigated a symmetry-aware surrogate-assisted approach for simulation-based testing of autonomous UAVs. By transforming obstacle configurations into a canonical flight frame, the approach enables a neural-network surrogate trained on a single reference mission to be reused across different flight missions without retraining. The surrogate estimates the minimum clearance distance between the UAV and surrounding obstacles, allowing potentially critical test cases to be identified before executing computationally expensive high-fidelity simulations.

The empirical evaluation demonstrated the effectiveness of the approach in generating critical test cases. Compared with the constrained-random generator used to produce the training data, the surrogate-assisted approach increased the critical test-case hit rate from approximately 10\% to 42\%. It also achieved competitive results against participants in the 2024–2026 UAV Testing Competitions, ranking first overall in 2024 and 2025 and second in 2026. However, the results also revealed limitations in transferring surrogate predictions across missions with different flight-path characteristics. Furthermore, although the approach achieved high input diversity, this did not consistently translate into high diversity in the resulting UAV trajectories.

Future work will investigate integrating the trained surrogate into search-based and evolutionary test-generation techniques. Since the surrogate can estimate obstacle clearance distance at a fraction of the computational cost of high-fidelity simulation, it offers the potential to evaluate large populations of candidate test cases efficiently and guide the search towards critical scenarios.
Another direction is to investigate neural-network-based generative approaches, including generative adversarial networks, for learning to produce critical obstacle configurations. The trained surrogate could serve as a computationally inexpensive evaluator to guide these generators towards failure-revealing scenarios. Finally, we plan to explore combining large language models with the surrogate to generate and evaluate critical test cases, exploiting the complementary capabilities of language models for scenario generation and neural-network surrogates for rapid prediction of test outcomes.
Future work will also investigate improving surrogate generalisation across missions by incorporating flight-context information and accounting for interactions with multiple flight-path segments.

\bibliographystyle{IEEEtran}
\bibliography{reference}

\end{document}